\documentclass[10.7pt,]{article}
\usepackage{hyperref}
\usepackage[letterpaper, margin=2.54cm, top=2.54cm]{geometry}
\usepackage[super,comma,sort&compress]{natbib}
\usepackage{lmodern}
\usepackage{authblk} % To add affiliations to authors
\usepackage{amssymb,amsmath}
\usepackage{wrapfig}
\usepackage{graphicx,grffile}
\usepackage[labelfont=bf,labelsep=period]{caption}
\usepackage{ifxetex,ifluatex}
\usepackage{fixltx2e} % provides \textsubscript
\ifnum 0\ifxetex 1\fi\ifluatex 1\fi=0 % if pdftex
  \usepackage[T1]{fontenc}
  \usepackage[utf8]{inputenc}
\else % if luatex or xelatex
  \ifxetex
    \usepackage{mathspec}
  \else
    \usepackage{fontspec}
  \fi
  \defaultfontfeatures{Ligatures=TeX,Scale=MatchLowercase}
\fi
\IfFileExists{upquote.sty}{\usepackage{upquote}}{}
\IfFileExists{microtype.sty}{%
	\usepackage{microtype}
	\UseMicrotypeSet[protrusion]{basicmath} % disable protrusion for tt fonts
}{}

\usepackage{lipsum} % for dummy text only REMOVE

\usepackage{sectsty}
\allsectionsfont{\fontsize{10}{10}\selectfont}
\subsectionfont{\itshape\bfseries\fontsize{10}{10}\selectfont}
\subsubsectionfont{\normalfont\itshape}

\makeatletter
\renewcommand\subsubsection{\@startsection{subsubsection}{3}{\z@}%
	{-3.25ex\@plus -1ex \@minus -.2ex}%
    {-1.5ex \@plus -.2ex}% Formerly 1.5ex \@plus .2ex
    {\normalfont\itshape}}
\makeatother

\makeatletter % Reference list option change
\renewcommand\@biblabel[1]{#1.} % from [1] to 1
\makeatother %

\usepackage{etoolbox}
\makeatletter
\patchcmd{\@maketitle}{\LARGE}{\bfseries\fontsize{15}{16}\selectfont}{}{}
\makeatother

\makeatletter
\def\maxwidth{\ifdim\Gin@nat@width>\linewidth\linewidth\else\Gin@nat@width\fi}
\def\maxheight{\ifdim\Gin@nat@height>\textheight\textheight\else\Gin@nat@height\fi}
\makeatother

\setkeys{Gin}{width=\maxwidth,height=\maxheight,keepaspectratio}
\ifx\paragraph\undefined\else
\let\oldparagraph\paragraph
\renewcommand{\paragraph}[1]{\oldparagraph{#1}\mbox{}}
\fi
\ifx\subparagraph\undefined\else
\let\oldsubparagraph\subparagraph
\renewcommand{\subparagraph}[1]{\oldsubparagraph{#1}\mbox{}}
\fi
\hypersetup{
	unicode=true,
	pdftitle={Active Learning with Feature Matching for Clinical Named Entity Recognition},
	pdfauthor={Linh Le, Guido Zuccon, Gianluca Demartini},
	pdfkeywords={Active Learning, Named Entity Recognition},
	pdfborder={0 0 0},
	breaklinks=true
}
\usepackage{microtype,graphicx,subfigure,booktabs,amsmath,amssymb,amsthm,mathtools,multirow}
\usepackage{subfigure}
\usepackage{soul,color}
\definecolor{ForestGreen}{rgb}{0.13, 0.55, 0.13}

\title{\vspace{-2em} Leveraging Semantic Type Dependencies for Clinical Named Entity Recognition}

\author[1]{\bf\fontsize{13}{14}\selectfont Linh Le}
\author[1]{\bf\fontsize{13}{14}\selectfont Guido Zuccon}
\author[1]{\bf\fontsize{13}{14}\selectfont Gianluca Demartini}
\author[2]{ \bf\fontsize{13}{14}\selectfont Genghong Zhao}
\author[3]{ \bf\fontsize{13}{14}\selectfont Xia Zhang}
\affil[1]{\bf\fontsize{13}{14}\selectfont University of Queensland, Australia \par linh.le, g.zuccon, g.demartini@uq.edu.au }
\affil[2]{\bf\fontsize{13}{14}\selectfont Neusoft Research of Intelligent Healthcare Technology, Co. Ltd., Shenyang, China \par zhaogenghong@neusoft.com}
\affil[3]{\bf\fontsize{13}{14}\selectfont Neusoft Corporation, Shenyang, China \par  zhangx@neusoft.com}
\date{} % add no date (by default date is added)
\usepackage{graphicx,amsmath,subfigure,multirow,adjustbox}
\usepackage{arydshln}
\begin{document}
\maketitle
\vspace{1em} %separation between the affiliations and abstract
%==============================

\section{Abstract}
Previous work on clinical relation extraction from free-text sentences leveraged information about semantic types from clinical knowledge bases as a part of entity representations. In this paper, we exploit additional evidence by also making use of \textit{domain-specific semantic type dependencies}. We encode the relation between a span of tokens matching a Unified Medical Language System (UMLS) concept and other tokens in the sentence. We implement our method and compare against different named entity recognition (NER) architectures (i.e., BiLSTM-CRF and BiLSTM-GCN-CRF) using different pre-trained clinical embeddings (i.e., BERT, BioBERT, UMLSBert). Our experimental results on clinical datasets show that in some cases NER effectiveness can be significantly improved by making use of domain-specific semantic type dependencies. Our work is also the first study generating a matrix encoding to make use of more than three dependencies in one pass for the NER task.

\section{Introduction}

The identification of clinical entities in unstructured text aims to extract detailed patients’ concepts such as diseases, diagnoses, and treatments. An emerging line of research for creating effective clinical NER approaches focuses on world-level representations by alternating or concatenating tokens $x_i$ from a sequence $X$ ($\{x_1, x_2, ..., x_n\}$)  with the best matching concept and semantic type of that concept \cite{amia/Wu18,jmri/Jiang19}. In this paper, we embrace that direction and we explore the impact of information about the dependency between words on the clinical NER methods. Previous work leveraging dependency information showed that relations between words could be represented by their distance and their syntactic relations \cite{amia/Wu18,emnlp/Jie19}. For example, syntactic relations may indicate the presence of some named entities  (e.g., the pattern \{compound, object\} may match the clinical treatment ``stulz neuro'').

\begin{figure*}[t]
  \centering
  \includegraphics[width=1\textwidth]{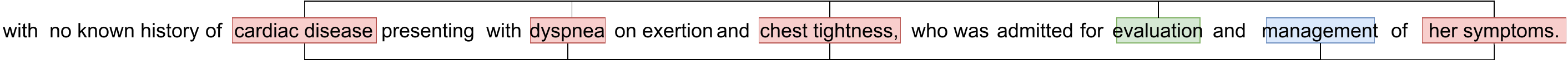}
  \caption{An example of a sentence annotated with named entities, short dependencies and long dependencies with clinical knowledge exploitation. There are 3 types of entities and two types of dependencies in this example. The entity type `disorder/problem'  is highlighted using red, `test' with green,  and `treatment' with blue. Long dependencies are relations between tokens not in the same clinical concept, which are \textit{treats} (between a `disorder/problem' entity and a `treatment' entity) and \textit{diagnoses} (between a `treatment/disorder' entity and a 'test' entity). Examples of a \textit{treats} dependency are (``cardiac disease'', management''), (``dyspnea, ``management''), (``chest tightness'', management''), and (``symptoms, ``management''). Examples of a  \textit{diagnoses} dependency are (``cardiac disease, ``evaluation), (``dyspnea, ``evaluation), (``chest tightness, ``evaluation), and (``symptoms'', ``evaluation''). Short dependencies are relations between tokens in clinical concepts, i.e.,  (``cardiac'', ``disease'') and (``chest'', ``tightness''). }\label{fig:dependency_example}
\end{figure*}
Inspired by the idea of leveraging more information about the dependencies between tokens, we propose a multiple semantic type dependency method applied to the clinical domain 
that captures both \textit{long-distance} and \textit{short-distance} interactions.
The long-distance interactions (long dependency) are obtained by considering clinical relationships between concepts of four semantic type groups (i.e, disorder, symptom, treatment, test) as done by Van Der Vegt et al. \cite{amia/VegtZK19}. For example, Figure \ref{fig:dependency_example} illustrates the long-distance interaction between the concepts \{``dyspnea''\} and \{``evaluation''\} by means of the \textit{diagnoses} relationship.
% Another long-distance interaction could be identified as the ``treats'' relationship between  \{``chest tightness'', ``management''\}.
Short-distance interactions (short dependency) are obtained by leveraging clinical concepts detected by an unsupervised NER approach (e.g., QuickUMLS). For example, ``cardiac'' has a short-distance interaction with ``disease'' via the ``cardiac disease'' concept detected by QuickUMLS.
These examples show how domain-specific semantic type information may be beneficial to perform NER more effectively. By guiding the machine learner using relations between tokens, the learner can pay more attention to the tokens that are more likely to represent potential clinical entities. To our knowledge, our work is the first to exploit both types of token dependencies.
Different from previous works for semantic type integration \cite{jbi/20/luo, ijcai/18/wang,emnlp/18/sun}, our work does not focus on training two tasks (NER and relation extraction) simultaneously but instead it exploits multiple relation types between two tokens in one pass for the NER task to reduce the training cost (i.e., both time and memory). The research question we address in this work is whether incorporating multiple dependencies without limiting the number of dependencies into natural language representations is beneficial for improving NER effectiveness.

\section{Background}
Previous work has shown that NER models in the clinical domain perform better when making use of external domain knowledge. For example, Wu et al. \cite{amia/Wu18} integrated a semantic type and lexical matching module into a BiLSTM-CRF architecture combined with Glove embedding \cite{emnlp/Pennington14}. This method surpassed Glove+BiLSTM-CRF model performance by 0.35 of F1 score. 
Lee et al. \cite{bioinformatics/Lee20} introduced a BERT-based model which was pre-trained on a biomedical corpus (i.e., PubMed abstracts and PubMed full-text articles) and outperformed a BERT-based linear model by up to 2.5 of F1 score.
Michalopoulos et al. \cite{naacl/Michalopoulos21} proposed UmlsBERT to connect word mapping to the same concepts in UMLS \cite{nar/boren} and exploit the semantic type knowledge in UMLS to augment input embeddings. UmlsBERT outperformed BioBERT by 1.37 of F1 score on the i2b2 V/A 2010 dataset. Despite such improvements show that the benefit of using domain knowledge in the NER task, the interactions between different tokens are not considered by the previous work on clinical NER, contrary to what has been attempted in general domain NER where syntactic relations are leveraged \cite{emnlp/Jie19}.
In our work, we analyze the impact of injecting clinical knowledge in the form of both \textit{short dependencies} and \textit{long dependencies}. Short dependencies are relations between tokens in a text span detected by UMLS concepts, and long dependencies are relations detected by their semantic type relationships in a sentence.
The work of Jie et al. \cite{emnlp/Jie19} is most similar to ours. Instead of having only one dependency between two tokens as them, we leverage multiple types of dependencies between two tokens with no constraint on the number of relations used. 

\section{Methods}
Next, we discuss the observations that lead us to the design of the proposed method (see Section \textit{Motivation}), present some tools used to extract clinical concepts, and link them to the external domain knowledge-base UMLS (see Section \textit{Tool Performance Analysis}), intuitively introduce our method (see Section \textit{Learning UMLS Word Relation}), and provide more details on the proposed method formulation relative to two different architectures: BiLSTM-CRF (see Section \textit{BiLSTM-CRF with Semantic Type Dependency}) and Graph Convolutional Networks (see Section \textit{Multiple Graph Convolutional Network Encoder}).

\subsection{Motivation} \label{Section:DataAnalysis}
Our idea of modeling token dependencies is based on the fact that `problem' entities and related-problem dependencies are common in clinical datasets. For example, the `problem' entity type in the i2b2/VA 2010 accounts for 49.16\% and 47.51\% of the training and test set respectively. The entities with related-problem dependencies are approximately $60\%$ and $48\%$ of the training and test set respectively (see Table \ref{table:i2b2EntityRelationStats} for details).
\begin{table*}[t]
\center
\caption{Statistics of entities with their relationships in the i2b2 V/A 2010 test dataset.}
\label{table:i2b2EntityRelationStats}
\begin{tabular}{lcc}
\hline
\textbf{Dependency} & \textbf{Training(\%)}& \textbf{Test(\%)}\\
\hline
problem entities not in any relationships  &14.85&13.28\\
treatment entities not in any relationships&10.51&10.51\\
test entities not in any relationships     &6.52& 6.68 \\ \hline
problem entities in treats relationships   &12.20&13.00 \\
problem entities in diagnoses relationships&14.53&15.53 \\ \hline
treatment entities in treats relationships &11.70&12.32 \\ \hline
% treatment entities in treatment-test relationships  & 1206 \\ \hline
test entities in diagnoses relationships   &9.40&10.10\\ 
% test entities in treatment-test relationships  & 1206 \\
\end{tabular}
\end{table*}
This observation leads us to consider the benefits of encoding these dependencies into the word embedding to improve entity recognition and in particular to reduce the number of false-positive predictions. For example, in Figure \ref{fig:dependency_example}, ``Evaluation'' is predicted as a `test' entity type because of two signals: (1) the dependency with other `problem' entities such as ``cardiac disease'', ``dyspnea'', ``chest tightness'', ``symptoms'', and (2) the semantic type in the `treat' semantic group from MetaMAP. However, this information is not guaranteed to be always correct due to being extracted from imperfect UMLS tools such as QuickUMLS and MetaMAP. Our proposed method provides a way to design deep learning architectures that can leverage noisy matching tools effectively.

\subsection{Tool Performance Analysis} \label{Section:ToolAnalysis}
In this section, we explain the mechanism and analyse the performance of two popular clinical concept extractors, MetaMAP \cite{jamia/aronson} and QuickUMLS \cite{sigir/Soldaini16} (see Table \ref{table:QuickUMLS-MetaMAP-results}). In this work, we use MetaMAP 20 \footnote{\url{https://lhncbc.nlm.nih.gov/ii/tools/MetaMap/Docs/MM_2020_ReleaseNotes.pdf}} and QuickUMLS $1.4.0$ \footnote{\url{https://pypi.org/project/quickumls/1.4.0.post1/}} to extract concepts.
Both MetaMAP and QuickUMLS employ unsupervised algorithms to approximately match clinical concepts in the UMLS Metathesaurus and provide semantic type information by looking for these concepts' semantic types in UMLS. In detail, they use a parser to take candidate phrases and generate lexical variations from each one; then, a matching algorithm is employed to select best-fit concepts in UMLS. For example, they can extract clinical concepts such as ``cardiac disease'', ``evaluation'', etc. in the example of Figure \ref{fig:dependency_example}.

Semantic types corresponding to these concepts are Disease or Syndrome for ``cardiac disease'' and Diagnostic Procedure for ``evaluation''. To evaluate these tools' performance, we use the semantic group of a concept as returned by the tools, and categorise them in three groups following Van der Vegt \cite{amia/VegtZK19}: disorder, treatment, test, corresponding to three entity types (disorder/problem, treatment, test). The QuickUMLS extractor results in higher recall for the `Disorder' and `Treatment' entity type but in lower recall for `Test' entities.\par 
\begin{table*}[!hb]
\caption{Performance evaluation of QuickUMLS and MetaMAP concept extractors.} \label{table:QuickUMLS-MetaMAP-results}
\begin{tabular}{p{0.1\textwidth}p{0.05\textwidth}p{0.05\textwidth}p{0.05\textwidth}p{0.05\textwidth}p{0.05\textwidth}p{0.05\textwidth}|p{0.05\textwidth}p{0.05\textwidth}p{0.05\textwidth}p{0.05\textwidth}p{0.05\textwidth}p{0.05\textwidth}}
\hline
\multirow{3}{*}{\textbf{Class}} &\multicolumn{6}{c}{\textbf{QuickUMLS}}  &\multicolumn{6}{c}{\textbf{MetaMAP}}\\
\cline{2-13}
 &
 \multicolumn{3}{c}{\textbf{i2b2/VA 2010}} & \multicolumn{3}{c}{\textbf{ShARe/CLEF 2013}}&
 \multicolumn{3}{c}{\textbf{i2b2/VA 2010}} & \multicolumn{3}{c}{\textbf{ShARe/CLEF 2013}}\\
 \cline{2-13}
 &P&R&F1 &P&R&F1&P&R&F1 &P&R&F1 \\
\hline
Treatment  & 33.48 & 73.15 & 45.93 &--&--&--& 44.56 & 69.19 & 54.21  &--&--&--  \\
Disorder   & 44.62 & 96.26 & 60.98 & 41.01 & 82.66 & 54.82   & 39.34 & 94.25 & 55.51 & 35.83 & 81.03  &   49.68\\
Test       & 67.59 & 26.47 & 38.04 & -- & -- & -- & 58.30 & 37.38 & 45.55 & -- & -- & --\\
\hline
\end{tabular}
\end{table*}
Based on our interest in leveraging all potential clinical dependencies between tokens, we select QuickUMLS as a component in our architecture because of this higher recall level. It should be noted that we use UMLS 2020AA version to identify UMLS concepts.

\subsection{Learning UMLS Word Relations}\label{sec:method_intuitition}

\begin{figure*}[!h]
  \centering
  \includegraphics[width=1\textwidth]{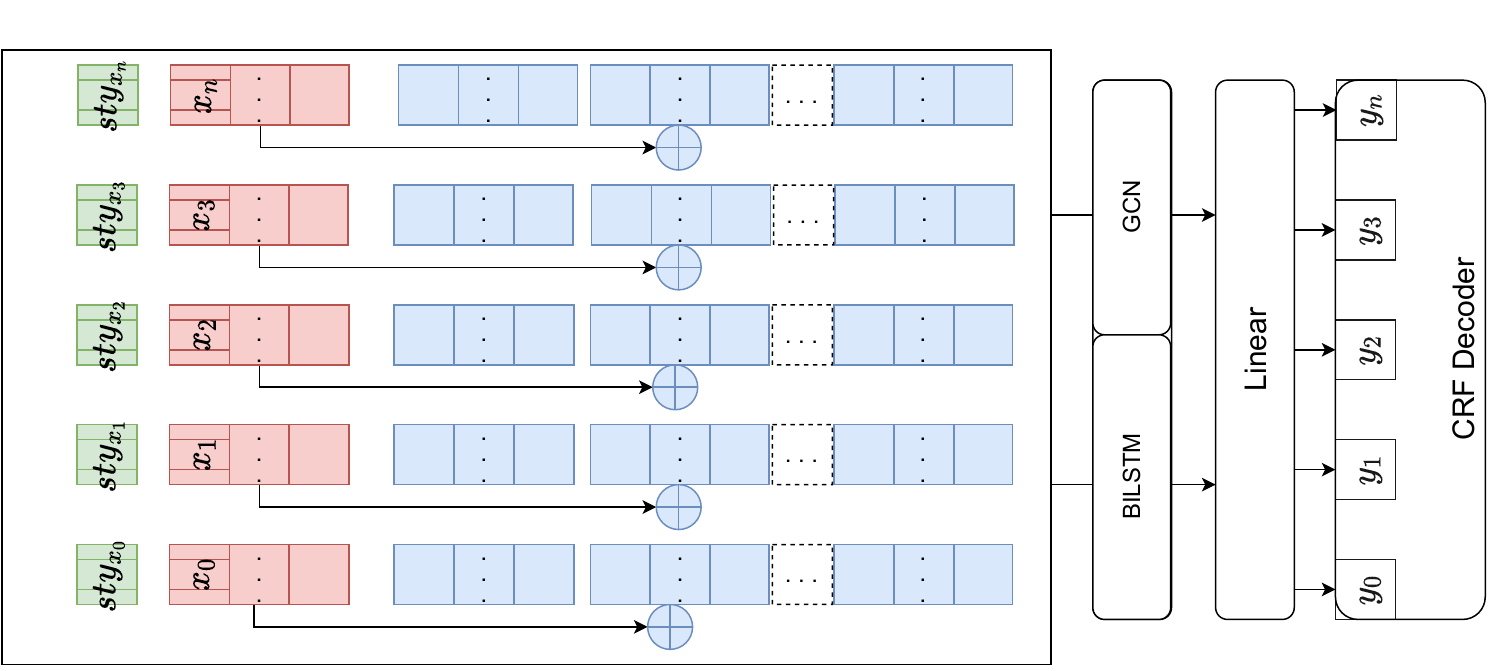}
  \caption{A NER system architecture utilizing multiple relationships between one token and others in a sequence. The blue components represent the word embedding of multiple tokens having a relationship with a token $x_i$ at position $i$ in a text sequence. Using the example in Figure \ref{fig:dependency_example}: the token ``disease'' in the concept ``cardiac disease'' is the red component in this architecture and other tokens such as ``evaluation'' and ``management''  are the blue components which are concatenated with token ``disease'' through different relation types such as ``treats'' and ``diagnoses''. Finally, the combination of word embeddings from these relationships is concatenated with the label encoding of the semantic type for the token at position $i$, which is the green component.}\label{fig:ner_architecture}
\end{figure*}

% The model we propose is most similar to dependency-guided NER~\cite{emnlp/Jie19}. %already stated
We use character embeddings, word embeddings, and sentence dependency encoding as input, and feed the concatenation of these into a BiLSTM and finally to the CRF layer for the classification (see Figure \ref{fig:ner_architecture} for more details). Instead of only using the syntactic relation between words, we also exploit the semantic type information available in domain-specific knowledge bases.
We obtain concepts and their corresponding semantic types using QuickUMLS.
To reduce the number of relations between tokens, we use the semantic type of a concept to establish their domain dependencies. Each token has multiple domain parent tokens to depend on. 
The structure dependencies of a sentence can be represented as a tree with a single token as the root (i.e., sentence structure tree). When modeling concepts, we can represent the tokens that belong to a concept (span) as a tree (i.e., concept tree). In this case, we use the last token of the span as the root, and all other tokens as sibling leaves. Because each concept is represented by a tree, the concept-based representation of a sentence results in multiple trees, each with its own root. 
% \gd{add example, or refer to example}
% 
For example, in Figure \ref{fig:dependency_example}, we have two root nodes (i.e., disease and evaluation) corresponding to two UMLS concepts (cardiac disease and evaluation). A token is then represented by the concatenation of three embeddings: its own embedding and the embedding of the parent token according to the domain concept tree and the semantic type label embedding. 
The multiple semantic type dependency relation embedding is initialized with the dimension $(r,768)$  where $r$ is the configured relation number ($max_r=n$, where n is the sequence length) and every position in the relation embedding is initialized to $0$ apart from the positions that correspond to the parent tokens in the concept tree (there may be multiple parent tokens), which are set to 1.
However, this embedding may become very big when we have more than two relations. Thus, tokens reduce the dimension of the relation embedding by averaging over all other related tokens embedding (see Eq. \ref{Equation:MultiRelation1Pass}) where $w_{d}$ denotes the word dependency embedding at position $i$ and is aggregated by an average function of other related clinical tokens $w_{r_j}$, and $m$ is the number of relations:
\begin{equation}
    w_{d} =\frac{1}{m}\times \sum_{j=0}^{m}{w_{r_j}}
    \label{Equation:MultiRelation1Pass}
\end{equation}

\noindent The semantic type label embedding is initialized to $0$ apart from the positions that correspond to a concept with identified semantic type in QuickUMLS (e.g., $1$ if \textit{Sign or Symptom}, $2$ if \textit{Finding}, etc.)

\subsection{BiLSTM-CRF with Semantic Type Dependency} \label{sec:bilstm_umls}
Similar to Lample et al. \cite{naacl/Lample15}, Reimers et al. \cite{emnlp/Reimers17}, we concatenate word embeddings with character-based embeddings and concatenate each word representation with its parents' word  
embedding. Differently from Jie et al. \cite{emnlp/Jie19}, we substitute a word embedding with unlimited parent dependencies. In detail, $w_i$ is the representation of the word $x_i$. The parent word $w_d$ consists of multiple related word embeddings and is aggregated via sum-pooling over the hidden states of the corresponding tokens in the last pretrained-model layer (see Eq. \ref{Equation:MultiRelation1Pass}).
We then concatenate $w_i$ and $w_d$ with $v_{d_r}$ - the embedding representation of the semantic types - and pass it to a BiLSTM, linear classifier with a CRF decoder to predict the token class.
We formulate the representation of the token at position $i$ as follows: $u_i=[w_i;w_d;v_{d_r}]$.
The above representation captures both the short-distance interaction short dependency and the long-distance interaction among tokens.

\subsection{Multiple Graph Convolutional Network (GCN) Encoder}\label{sec:gcn_umls}
We follow Kipf et al. \cite{iclr/Kipf16} and Zhang \cite{acl/Zhang18} to create a GCN-based method for NER (BiLSTM-GCN). To deal with multiple types of relations, we implement multiple GCNs and concatenate the embedding representations of relations between tokens identified through sentence dependency and semantic type dependency. We also use a CRF layer as the final layer for the NER task as in previous work \cite{jmlr/Collobert11, naacl/Lample15,acl/ChiuNichols16}. 
Given a graph $G=(V,E)$, we have a feature vector $x_v \in R^d$ for each node $v \in V$. The  representation $h_{v_j}$ for node $v_j$ is formulated as follows:
\begin{equation*}
\begin{split}
    h_{v_j} &= ReLU(\sum_{v_i \in N(v_j)}W_{D(i,j)}X_{v_i}+b_{D(i,j)})
\end{split}
\end{equation*}
where $N(v_j)$ is the set of the nodes directly connected to $v_j$, $D(i,j)$ indicates an edge between node $v_i$ and $v_j$ in the clinical dependency relation structure.
$W$, $b$, and $ReLU$ are the weight matrix, bias parameters, and the rectifier linear unit function of the model respectively.

\section{Experiments}
\subsection{Datasets}
The 2010 i2b2/VA Challenge dataset \cite{jamia/Uzuner11} is a collection of clinical notes with annotations for three entity types: `problem', `test', and `treatment'. It contains 14,504 sentences and 15,179 entities in the training set, 1809 sentences and 1,340 entities in the validation set, 27,624 sentences, and 31,161 entities in the test set. The CLEF 2013 dataset \cite{clef/Suominen13} identifies only one entity type, `disorder', and it contains 10,171 sentences in the training set and 9,273 sentences in the test set.

\subsection{Evaluation Metrics}
We use the precision (P), recall (R), F1-score (F1) metrics to evaluate the effectiveness of NER models \cite{naacl/Lample15,emnlp/Jie19}.
We compute the micro-average to weigh equally each sample and avoid the influence of class imbalance (i.e., there could be more examples of one class than of other classes). We compare methods' performance using significance tests. Specifically, we employ a bootstrap significance test because it can be used with all three metrics (P, R, F1) \cite{acl/Dror18}. We also use sample size=1'000 (>200, which is recommended in  previous work  \cite{acl/Sogaard14}). 

\subsection{Baselines and Methods}
Besides a linear model (labelled \textit{linear}), we implement two BiLSTM-CRF-based NER  methods for comparison: the BiLSTM-CRF \cite{naacl/Lample15} (\textit{Bi-CRF})  and the Dependency-Guided BiLSTM-CRF \cite{emnlp/Jie19} (\textit{dglstm}). 
We also use BiLSTM-CRF as a component in our methodology as described in Sections ``BiLSTM-CRF with Semantic Type Dependency'' and ``Multiple Graph Convolutional Network (GCN) Encoder''. We thus obtain our proposed methods which we label \textit{our Bi-CRF} for BiLSTM-CRF and \textit{our-GCN} for GCN. Our method exploits initial information extracted using QuickUMLS. To reduce the number of non-related entity tokens, we only extract concepts from 18 semantic types that are useful to support disorder-centric, clinical relationships as identified by previous work (e.g., symptoms, diagnostic tests, diagnoses, and treatments) \cite{amia/VegtZK19}.

\subsection{Experimental Setting}
 We use three pre-trained models as our word embeddings to consider our method's effectiveness on both non-clinical language model BERT \cite{naacl/Devlin19}, and the biomedical language models BioBERT \cite{bioinformatics/Lee20} and UMLSBert \cite{naacl/Michalopoulos21}. We implement our model on a Linux server with 4 CPUs, one GeForce RTX 2080 Ti GPU with $12$ GB of memory, a learning rate of 0.01, batch-size of 24, L2 regularisation parameter is $1e-8$. 
 
\section{Results and Discussion}
\subsection{Result Comparison on ShARe/CLEF 2013}
The results in Table \ref{table:CLEFResults} show that our model always achieves the best F1 across the considered methods (except for BERT+our Bi-CRF). Specifically, the highest F1 is achieved when using the BiLSTM-CRF architecture with UMLSBert embedding.
\begin{table*}[!h]
\centering
\caption{Micro-average performance of different models on the disorder entity type of ShARe/CLEF 2013 and i2b2/VA 2010 datasets. Bold values indicate higher results for our method, (*) to denote statistically significant difference between the higher results compared to the linear model with p<0.05, (+) to denote statistically significant difference between the higher results compared to the best baseline model with p<0.05.}
\begin{tabular}{lccc|cccccccc}
\hline
\multirow{3}{*}{\textbf{Model}}&\multicolumn{3}{c}{\textbf{ShARe/CLEF 2013}}&\multicolumn{6}{c}{\textbf{i2b2 V/A 2010}}\\
\cline{2-10}
&\multicolumn{3}{c}{\textbf{Test dataset}}&\multicolumn{3}{c}{\textbf{Test dataset}} & \multicolumn{3}{c}{\textbf{Validation dataset}} \\
\cline{2-10}
&\textbf{P}& \textbf{R}& \textbf{F1}&\textbf{P}& \textbf{R}& \textbf{F1}&\textbf{P}&\textbf{R}&\textbf{F1}\\
\hline
\hline
BERT+linear       &75.79&74.38&76.01&83.63&85.84&84.72&83.42&87.23&85.28\\
BERT+Bi-CRF       &78.10&71.14&74.46&82.76 & 84.52 & 83.63 & 82.71 & 86.11 & 84.37 \\ 
BERT+dglstm       &75.14&71.32&73.18&83.78 & 85.15 & 84.46 & 84.23 & \textbf{88.13} & \textbf{86.13}\\
BERT+our-GCN      & \textbf{78.11} & 73.97 & 75.98 &83.43 & 85.56 & 84.48 &82.83 & 86.86 & 84.79\\ 
BERT+our Bi-CRF   &\textbf{78.27}{+}& \textbf{75.64} &\textbf{76.93}{+}&\textbf{84.15} & \textbf{86.10} & \textbf{85.12}& \textbf{84.48} & 87.83 & 86.12\\ \hline
BioBERT+linear    &79.92&76.77&78.31&86.62 & 87.83 & 87.22&85.15 & 88.65 & 86.86 \\
BioBERT+Bi-CRF    &80.19&76.60&78.35& 86.02 & 87.19 & 86.60&84.90 & 88.57 & 86.70\\
BioBERT+dglstm    &79.93&76.57&78.2&86.31 & \textbf{87.80} & 87.05&85.03 & 88.20 & 86.58\\
BioBERT+our-GCN   & 79.86 & \textbf{77.65} &\textbf{ 78.74} &\textbf{86.61} & 87.64 & 87.12 &\textbf{86.25} & \textbf{89.02} & 87.61\\
BioBERT+our Bi-CRF      &\textbf{80.64}&\textbf{77.53}&\textbf{79.06}& 86.60 & 87.70 & \textbf{87.15}& 85.00 & 88.42 & 86.68\\ \hline
UmlsBERT+linear   &80.55&79.58&80.06& 87.84 & 89.35 & 88.59& 85.65 & 89.17 & 87.38\\ 
UmlsBERT+Bi-CRF   &80.57&79.34&79.95&87.34 & 89.02 & 88.17& 86.06 & 89.47 & 87.73  \\ 
UmlsBERT+dglstm   &81.03&79.44&80.22& 87.72 & 88.79 & 88.25& 86.84 & 89.10 & 87.96\\
UmlsBERT+our-GCN  & 81.12 & \textbf{79.89} & \textbf{80.50} & 87.60 & 88.99 & 88.29& 86.43 & 88.95 & 87.67\\ 
UmlsBERT+our Bi-CRF      &\textbf{82.40}{*}&\textbf{80.06}&\textbf{81.21}{*}& \textbf{87.96} & \textbf{89.68} & \textbf{88.81} & \textbf{87.04} & \textbf{90.81} & \textbf{88.89}\\ \hline
\end{tabular}
\label{table:CLEFResults}
\end{table*}

%on the ShARe/CLEF 2013 dataset by outperforming other baselines by $0.74\%-1.26\%$ using BiLSTM-CRF architecture with UMLSBert embedding. 
The advantage of our method is most evident when looking at precision, with increases of $1.28-1.85$. This improvement is due to the reduction of false-positive predictions. Our method gives the machine learner more constraints and evidence to predict why a specific token of a sequence belongs to an entity class. For example, in Figure \ref{fig:dependency_example}, the machine learner predicts that the ``cardiac'' token belongs to the `treatment' class by understanding the relationship between this token and the adjacent token ``disease'' and the relationships between this token and other tokens such as ``evaluation'' and ``management''. 
Using BERT embeddings, our method improves precision by $0.17-3.13$ points and recall by $1.26-4.5$. It also leads to a slight improvement in F1 compared to the best baseline ($0.93$). When BioBERT is used, our BiLSTM-CRF improves precision between $0.45-0.72$ points,  recall by $0.76-0.96$, and leads to a slight improvement in F1 ($0.75-0.85$); however, we do not observe a statistically significant differences as compared to baseline effectiveness ($p>0.05$). We also observe a slight improvement of F1 without a statistically significant difference when using a GCN architecture.

\subsection{Result comparison on i2b2/VA 2010}
Table \ref{table:CLEFResults} reports the results obtained on i2b2/VA 2010 across all classes (micro-average). The highest effectiveness is obtained by our method with the Bi-CRF architecture and UmlsBERT embedding, and higher improvements are achieved in precision, rather than recall.
However, improvements are more moderate when using other pre-trained word embeddings such as BioBERT and BERT on both test and validation ($<0.4\%$). 
\begin{table*}[!ht]
\centering
\caption{Performance of different model on i2b2/VA 2010 dataset and finetuned with pre-trained UmlsBERT model.}\label{table:i2b2UmlsBERTresults}
\begin{tabular}{l|c|ccc|ccc}
\hline
\multirow{2}{*}{\textbf{Model}}&\multirow{2}{*}{\textbf{Class}} &\multicolumn{3}{c}{\textbf{Test dataset}} & \multicolumn{3}{c}{\textbf{Validation dataset}} \\
\cline{3-8}
&&\textbf{P}& \textbf{R}& \textbf{F1}&\textbf{P}&\textbf{R}&\textbf{F1}\\
\hline
\multirow{3}{*}{ \adjustbox{stack=ll,width=1.0\textwidth}{UmlsBERT\\+linear}}
&problem& 87.75 & 90.04 & 88.88 & 84.88 & 90.68 & 87.68 \\
&treatment&88.14 & 89.26 & 88.70 & 86.78 & 87.20 & 86.99\\
&test&87.65 & 88.51 & 88.08 & 85.58 & 89.22 & 87.36 \\\cdashline{1-8}
% &avg& 87.84 & 89.35 & 88.59 & 27734 & 3841 & 3304 & 85.67 & 89.17 & 87.37 & 1194 & 200 & 145\\ 
\multirow{3}{*}{\adjustbox{stack=ll,width=1.0\textwidth}{UmlsBERT\\+Bi-CRF}}
&problem&87.51 & 89.32 & 88.40 & 84.95 & 90.11 & 87.45 \\
&treatment&87.40 & 88.64 & 88.02  & 86.94 & 88.41 & 87.66 \\
&test& 87.05 & 89.00 & 88.02 &86.68 & 89.72 & 88.18\\\cdashline{1-8} 
% &avg/total &87.34 & 89.02 & 88.17 & 27631 & 4005 & 3407 & 86.08 & 89.47 & 87.73 & 1198 & 194 & 141 \\
\multirow{3}{*}{\adjustbox{stack=ll,width=1.0\textwidth}{UmlsBERT\\+dglstm}}
&problem& 87.69 & 89.25 & 88.46 & 85.48 & 90.68 & 88.01 \\
&treatment&88.58 & 88.15 & 88.36  & 89.30 & 86.51 & 87.88 \\
&test&86.92 & 88.79 & 87.84  & 86.27 & 89.72 & 87.96 \\\hline
% &avg/total&87.73 & 88.79 & 88.25 & 27558 & 3858 & 3481& 86.90 & 89.10 & 87.95 & 1194 & 181 & 146\\
\multirow{3}{*}{\adjustbox{stack=ll,width=1.0\textwidth}{UmlsBERT\\+our GCN }}
&problem&\textbf{87.88} & 89.27 & 88.57 &\textbf{86.21} & 90.30 & \textbf{88.21} \\
&treatment& 88.03 & 88.72 & 88.38 &86.20 & 85.99 & 86.09 \\
&test&86.81 & \textbf{88.87} & 87.83 & \textbf{86.96}{*} & \textbf{90.23} & \textbf{88.56}{*} \\\cdashline{1-8}
% &avg/total& 87.61 & 88.99 & 88.29 & 27620 & 3908 & 3418& 86.43 & 88.95 & 87.66 & 1191 & 187 & 14 \\
\multirow{3}{*}{\adjustbox{stack=ll,width=1.0\textwidth}{UmlsBERT\\+our Bi-CRF}}
&problem& \textbf{88.07} & 89.91 & \textbf{88.98} & \textbf{86.98}{+}& \textbf{91.44}{+} & \textbf{89.16}{+} \\
&treatment& \textbf{88.64} & \textbf{89.56} & \textbf{89.10} &87.41 & \textbf{88.89}{*} & \textbf{88.14}{*} \\
&test& 87.13 & \textbf{89.47} & \textbf{88.28}&\textbf{86.76}{+} & \textbf{91.98}{+} & \textbf{89.29}{+}
% &avg/total  & 87.96 & 89.68 & 88.81 & 27834 & 3811 & 3204&87.05\textbf{} & \textbf{90.81} & \textbf{88.88} & \textbf{1216} & \textbf{181} & \textbf{123}\\
\end{tabular}
\label{table:i2b2OurResults}
\end{table*}

With more than one entity class available in the i2b2/VA 2010 dataset, we study our method performance on each entity type separately; for space limitations, we report these detailed results for the best embedding model only (UmlsBERT, Table \ref{table:i2b2UmlsBERTresults}).
Among all baseline methods, dglstm provides the highest F1 and P on the three entity types, and the highest R for two out of three entity types (i.e., problem, test). The linear model achieves the best recall for the `problem' entity type, while BiLSTM-CRF achieves the best recall for `treatment' and `test' entities.
Our methods are overall better than the baselines, and our Bi-CRF architecture is always better than our GCN one (except for the ``test'' entity, where GCN has minor gains over Bi-CRF).
Similar to what is observed on ShARe/CLEF 2013, our method increases precision more than recall. The increment is $1.14-1.74$  and $0.38-0.05$ in precision and recall respectively for the `problem' class, $1.49-2.9$ and $1.45-2.14$ for the `treatment' class and $0.91-2.31$ and $1.26-1.76$ for the `test' class in the validation dataset. We observe a minor increment without statistically significant difference over the best baseline by $0.32-0.56$ in precision for the `problem' class, $0.06-1.24$, and $0.3-1.41$ in precision and recall for the `treatment' class, $0.47-0.96$ in recall for the `test' class. Class-based improvements are quite similar when BERT and BioBERT embeddings are used. Differences are however observed, in particular, for our GCN-based method with BioBERT embeddings as it achieves better effectiveness for `test' entities in the validation set compared to all baselines ($p<0.05$).

\subsection{Effect of Multiple Dependency Embedding}
To better understand how our method achieves improvements in precision, we consider the results on i2b2/VA 2010 and construct the confusion matrix between our method and the best baseline when using the UmlsBERT embeddings (see Table \ref{table:best_CFmatrix_valid} for the validation set and Table \ref{table:best_CFmatrix_test} for the test set). Our method reduces the number of wrong predictions with 'O' label, by approximately $5\%-6.1\%$ on the test dataset and by up to $10.37\%$ on the validation dataset. 
\begin{table*}[!ht]
\center
\caption{Confusion matrix results for the two best models on the i2b2 V/A 2010 validation dataset with UmlsBERT embeddings.}
\label{table:best_CFmatrix_valid}
\begin{tabular}{p{0.15\textwidth}|p{0.1\textwidth}p{0.1\textwidth}|p{0.1\textwidth}p{0.1\textwidth}|p{0.1\textwidth}p{0.1\textwidth}|p{0.1\textwidth}}
\hline
\multirow{2}{*}{Best Baseline} &\multicolumn{2}{c}{Problem}&\multicolumn{2}{c}{Treatment}&\multicolumn{2}{c}{Test}&\multirow{2}{*}{O} \\
&Bproblem & Iproblem& Btreatment& Itreatment&Btest& Itest&\\
\hline
Bproblem&495&12&7&0&1&0&39\\
Iproblem&11&666&1&12&1&1&68\\\cdashline{1-8}
Btreatment&1&1&374&\textbf{6}&\textbf{0}&0&21\\
Itreatment&0&2&\textbf{7}&332&1&4&35\\\cdashline{1-8}
Btest&\textbf{0}&1&6&1&371&6&24\\
Itest&0&3&0&7&12&\textbf{294}&25\\\cdashline{1-8}
O&19&30&19&\textbf{25}&13&\textbf{11}&9834\\
\end{tabular}
\begin{tabular}{p{0.15\textwidth}|p{0.1\textwidth}p{0.1\textwidth}|p{0.1\textwidth}p{0.1\textwidth}|p{0.1\textwidth}p{0.1\textwidth}|p{0.1\textwidth}}
\hline
\multirow{2}{*}{Our method} &\multicolumn{2}{c}{Problem}&\multicolumn{2}{c}{Treatment}&\multicolumn{2}{c}{Test}&\multirow{2}{*}{O} \\
&Bproblem & Iproblem& Btreatment& Itreatment&Btest& Itest&\\
\hline
Bproblem&495&7&\textbf{4}&\textbf{0}&\textbf{0}&\textbf{0}&\textbf{38}\\
Iproblem&12&\textbf{672}&\textbf{0}&\textbf{7}&\textbf{0}&\textbf{0}&\textbf{60}\\\cdashline{1-8}
Btreatment&1&1&\textbf{377}&7&1&0&\textbf{21}\\
Itreatment&0&2&10&330&\textbf{0}&\textbf{2}&\textbf{25}\\\cdashline{1-8}
Btest&6&0&\textbf{4}&1&\textbf{376}&\textbf{4}&\textbf{20}\\
Itest&0&8&\textbf{0}&\textbf{5}&12&291&\textbf{26}\\\cdashline{1-8}
O&\textbf{12}&\textbf{25}&19&33&\textbf{10}&19&\textbf{9856}\\
\end{tabular}
\end{table*}

\begin{table*}
\center
\caption{Confusion matrix results for the two best models on the i2b2 V/A 2010 test dataset with UmlsBERT embeddings.}
\begin{tabular}{p{0.15\textwidth}|p{0.1\textwidth}p{0.1\textwidth}|p{0.1\textwidth}p{0.1\textwidth}|p{0.1\textwidth}p{0.1\textwidth}|p{0.1\textwidth}}
\hline
\label{table:best_CFmatrix_test}
\multirow{2}{*}{Best Baseline} &\multicolumn{2}{c}{Problem}&\multicolumn{2}{c}{Treatment}&\multicolumn{2}{c}{Test}&\multirow{2}{*}{O} \\
&Bproblem & Iproblem& Btreatment& Itreatment&Btest& Itest&\\
\hline
Bproblem&\textbf{11728}&\textbf{155}&55&8&48&5&693\\
Iproblem&215&16368&11&108&28&115&1487\\\cdashline{1-8} 
Btreatment&\textbf{50}&17&8476&152&64&6&470\\
Itreatment&\textbf{3}&\textbf{79}&\textbf{149}&6942&\textbf{0}&73&574\\\cdashline{1-8} 
Btest&\textbf{40}&\textbf{43}&\textbf{83}&3&8367&226&431\\
Itest&\textbf{0}&\textbf{83}&2&\textbf{69}&90&7040&495\\\cdashline{1-8} 
O&\textbf{537}&926&487&\textbf{631}&605&\textbf{537}&197725\\
\end{tabular}
\begin{tabular}{p{0.15\textwidth}|p{0.1\textwidth}p{0.1\textwidth}|p{0.1\textwidth}p{0.1\textwidth}|p{0.1\textwidth}p{0.1\textwidth}|p{0.1\textwidth}}
\hline
\multirow{2}{*}{Our method} &\multicolumn{2}{c}{Problem}&\multicolumn{2}{c}{Treatment}&\multicolumn{2}{c}{Test}&\multirow{2}{*}{O} \\
&Bproblem & Iproblem& Btreatment& Itreatment&Btest& Itest&\\
\hline
Bproblem&11718&156&\textbf{44}&\textbf{5}&\textbf{39}&\textbf{7}&\textbf{635}\\
Iproblem&\textbf{194}&16368&16&\textbf{84}&\textbf{30}&\textbf{90}&\textbf{1411}\\\cdashline{1-8} 
Btreatment&\textbf{56}&\textbf{12}&\textbf{8488}&134&\textbf{45}&\textbf{5}&\textbf{449}\\
Itreatment&6&92&156&6942&\textbf{1}&\textbf{57}&\textbf{536}\\\cdashline{1-8} 
Btest&50&52&100&\textbf{2}&\textbf{8452}&\textbf{210}&\textbf{423}\\
Itest&1&93&\textbf{1}&70&\textbf{75}&7033&\textbf{448}\\\cdashline{1-8} 
O&548&\textbf{900}&\textbf{458}&676&\textbf{560}&600&\textbf{197973}\\
\end{tabular}
\end{table*}

We also consider how the relationships \textit{treats} and \textit{diagnoses} increase precision. These relationships could reduce the number of incorrect predictions from class A to class B, where A and B are entities in those \textit{diagnoses} and \textit{treats} relationships. The incorrect predictions for `treatment' (ground truth) to `problem' (prediction) are reduced from 182 to 149 in the test dataset and of about a half in the validation set. The incorrect predictions for `test' to `problem' are reduced from 196 to 166 in the test dataset and no incorrect predictions in the validation set.
Due to the disorder-centric relations, our method lowers the number of errors made when incorrectly labeling a `test' or `treatment' entity as `problem', but it does not perform better for other situations.
Thus, for a disorder-centric dataset, the proposed method may be an effective alternative to incorporating domain knowledge into NER systems. This would be advantageous as this is less expensive than creating models pre-trained on extensive clinical corpora.

\section{Conclusions}
In this paper, we integrate external clinical knowledge in two different NER architectures (i.e., BiLSTM and GCN) through multiple token dependencies in a single pass. Our experimental evaluation results show that the proposed method achieves better NER effectiveness as compared to other baselines when combined with the BERT and UmlsBERT pre-trained models according to F1 on ShARe/CLEF 2013 and, in some cases, better than other baselines on i2b2/VA 2010. Our results confirm the benefits of using multiple dependency information for NER performed on clinical notes.

\section{Acknowledgments}
This research is supported by the National Key Research and Development Program of China No. 2020AAA0109400 and the Shenyang Science and Technology Plan Fund (No. 21-102-0-09).

%==============================
\bibliographystyle{vancouver}
\bibliography{literature}
%==============================

\end{document}